\documentclass{egpubl}
\usepackage{eg2021}
 
\Poster                 
\usepackage[T1]{fontenc}
\usepackage{dfadobe}  
\usepackage{cite}  
\BibtexOrBiblatex
\electronicVersion
\PrintedOrElectronic
\ifpdf \usepackage[pdftex]{graphicx} \pdfcompresslevel=9
\else \usepackage[dvips]{graphicx} \fi

\usepackage{egweblnk} 

\usepackage{times}
\usepackage{epsfig}
\usepackage{graphicx}
\usepackage{amsmath}
\usepackage{amssymb}
\usepackage{amsfonts}
\usepackage{amsbsy}
\usepackage{microtype}
\usepackage[ruled,vlined]{algorithm2e}
\usepackage{makecell}
\usepackage{multirow}
\usepackage{adjustbox}
\usepackage{bm}
\usepackage[dvipsnames]{xcolor}
\usepackage{hyperref}
\usepackage{wrapfig}
\DeclareSymbolFont{Xlargesymbols}{OMX}{cmex}{m}{n}
\DeclareMathSymbol{\Xsum}{\mathop}{Xlargesymbols}{80}

\begin{document}


\title{Generative Landmarks}
\author[D. Ferman \& G. Bharaj]
{\parbox{\textwidth}{\centering D. Ferman
        and G. Bharaj
        }
        \\
{\parbox{\textwidth}{\centering AI Foundation, USA}}}

%

 \teaser{
  \centering
  \includegraphics[width=\linewidth]{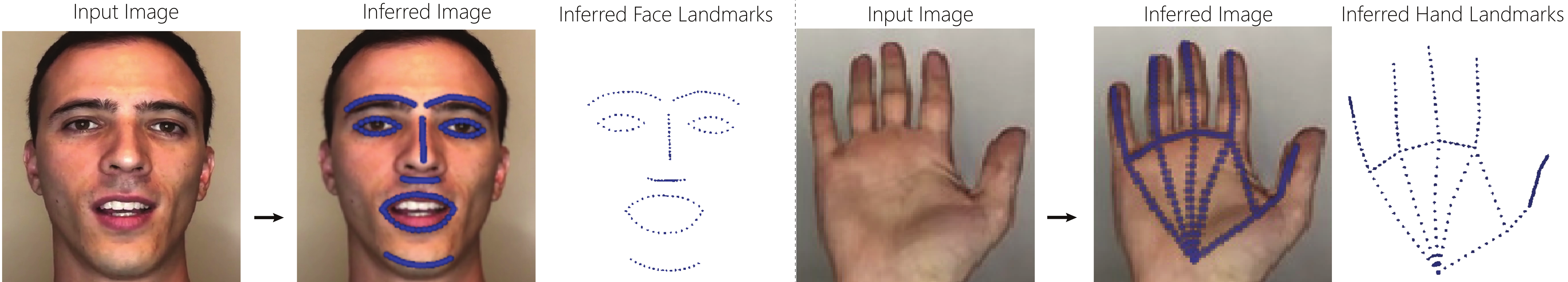}
  \caption{(Left and right) Unmarked input image, inferred image with markers, and inferred template landmarks.}
 \label{fig:teaser}
}
\maketitle
\begin{abstract}
    We propose a general purpose approach to detect landmarks with improved temporal consistency, and personalization. Most sparse landmark detection methods rely on laborious, manually labelled landmarks, where inconsistency in annotations over a temporal volume leads to sub-optimal landmark learning. Further, high-quality landmarks with personalization is often hard to achieve. We pose landmark detection as an image translation problem. We capture two sets of unpaired marked (with paint) and unmarked videos. We then use a generative adversarial network and cyclic consistency to predict deformations of landmark templates that simulate markers on unmarked images until these images are indistinguishable from ground-truth marked images. Our novel method does not rely on manually labelled priors, is temporally consistent, and image class agnostic -- face, and hand landmarks detection examples are shown.
\ccsdesc[500]{Computing methodologies~Interest point and salient region detections}
\ccsdesc[500]{Computing methodologies~Tracking}
\printccsdesc
\end{abstract}  
\section{Introduction}
\label{sec:Introduction}
\vspace{-5px}
Sparse landmarks detection is an important problem for face detection applications \cite{amos2016openface}, face tracking with landmarks alignment as a sub-task for 3D face model fitting~\cite{bickel2007multi, dib2021practical} or to guide video synthesis for faces ~\cite{wang2020facevid2vid}, other body parts~\cite{chan2019everybody}, among several others. These tasks rely on high-quality and temporally consistent landmarks; however, off-the-shelf landmark detection methods suffer from inconsistencies due to ambiguity in manual landmark annotations as well as temporal imperfections of frame-to-frame labeling, as landmarks are difficult to define precisely, see \cite{dong2018supervision,wu2018look} for a discussion. As a result, landmark detection models suffer from temporal jitters, and sub-optimal personalization. Wu et al.~\cite{wu2018look} approach this problem with a focus on the boundary, taking advantage of well defined face boundary lines along which the landmarks reside. Dong et. al.~\cite{dong2018supervision} note that frame-to-frame landmark detection should ideally resemble the presence of physical markers and present an approach that uses optical flow, and later a triangulation-based approach ~\cite{dong2020supervision} that exploits the temporal information inherent in video data.
\\
With the goal of temporally consistent and personalized landmark detection, we propose a method that involves capture of two sets of unpaired videos for a given body region: one in which semantically (e.g. eyes, nose, fingers, etc.) meaningful lines are visibly marked, and the other, unmarked. We predict the landmark deformations for a template, for each unmarked image and render them, such that, it resembles the marked images. Following Zhu et al.~\cite{zhu2017unpaired}, we pose this problem as an unpaired image translation problem, with an image generator network that translates from marked-to-unmarked images, while our novel landmark deformation prediction network performs the reverse translation. Thus, our method is capable of learning a set of predefined landmarks in an unsupervised fashion, circumventing the need for laborious and imprecise manual annotations, while providing landmarks that are inherently personalized and temporally stable. 
\begin{figure*}[t]
  \includegraphics[width=\linewidth]{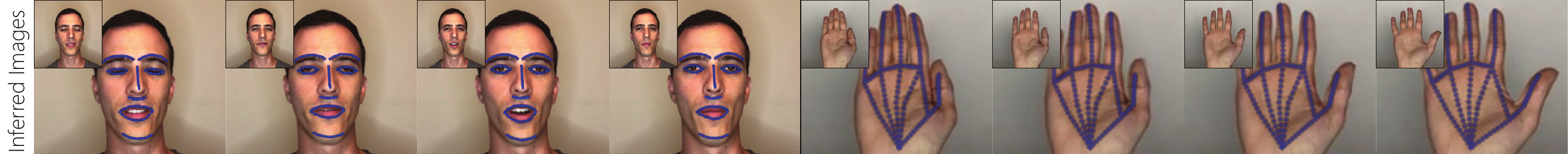}
  \label{fig:results}
  \vspace{-10px}
  \caption{(Left and right) Temporally consistent landmark predictions for faces and hands.}
  \vspace{-20px}
\end{figure*}
%
\section{Method: Generative Landmarks}
\label{sec:method}
\begin{figure}
  \includegraphics[width=\linewidth]{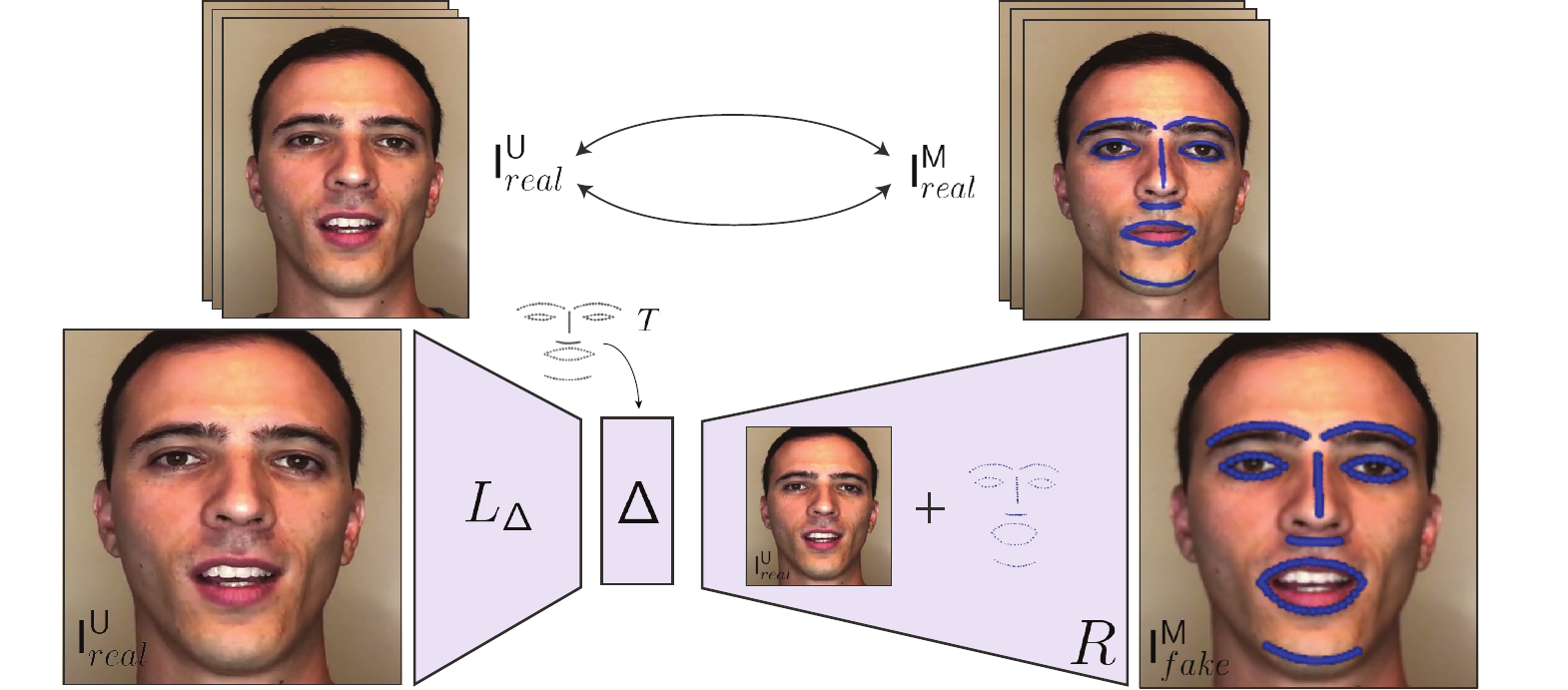}
   \caption{\label{fig:ex1}
        Top: Cyclic consistency between $\mathsf{I}^{U}_{real}$ and $\mathsf{I}^{M}_{real}$. Bottom: Unmarked image $\mathsf{I}^{U}_{real}$, is fed into the encoder, $L_\mathsf{\Delta}$, that gives landmark deformations $\mathsf{\Delta}$. The deformations are applied to template $\mathsf{T}$, and combined with $\mathsf{I}^{U}_{real}$ via our decoder -- differentiable renderer $R$. The resultant image is treated as a fake marked image $\mathsf{I}^{M}_{fake}$.}
\vspace{-30px}
\end{figure}

\vspace{-5px}
Given two unpaired image sets -- marked $\{\mathsf{I}^\mathsf{M}\}$ and unmarked $\{\mathsf{I}^\mathsf{U}\}$, our goal is to train a landmark deformation network, $L_{\mathsf{\Delta}}: \mathsf{I}^{\mathsf U} \rightarrow \mathsf{I}^{\mathsf M}$, that \textit{takes-in} unmarked images, $\mathsf{i}_i^{\mathsf M} \in \mathbb{R}^{3 \times H \times W}$ and predicts landmark deformations, learning from the marked images in an unsupervised fashion. Rather than predicting these landmarks directly, we use a template $\mathsf{T}\in \mathbb{R}^{N \times 2}$ with predefined spatial landmarks $t_i$, that form lines corresponding to the marked image set. In $L_{\mathsf{\Delta}}$, we intrinsically predict landmark deformations, $\mathsf{\Delta} \in \mathbb{R}^{N \times 2}$, that are applied as offsets to the template $\mathsf{T}$ (below) and rendered onto the unmarked images.
\begin{wrapfigure}{l}{0.3\linewidth}
\vspace{-10px}
    \centering\includegraphics[width=1.2\linewidth]{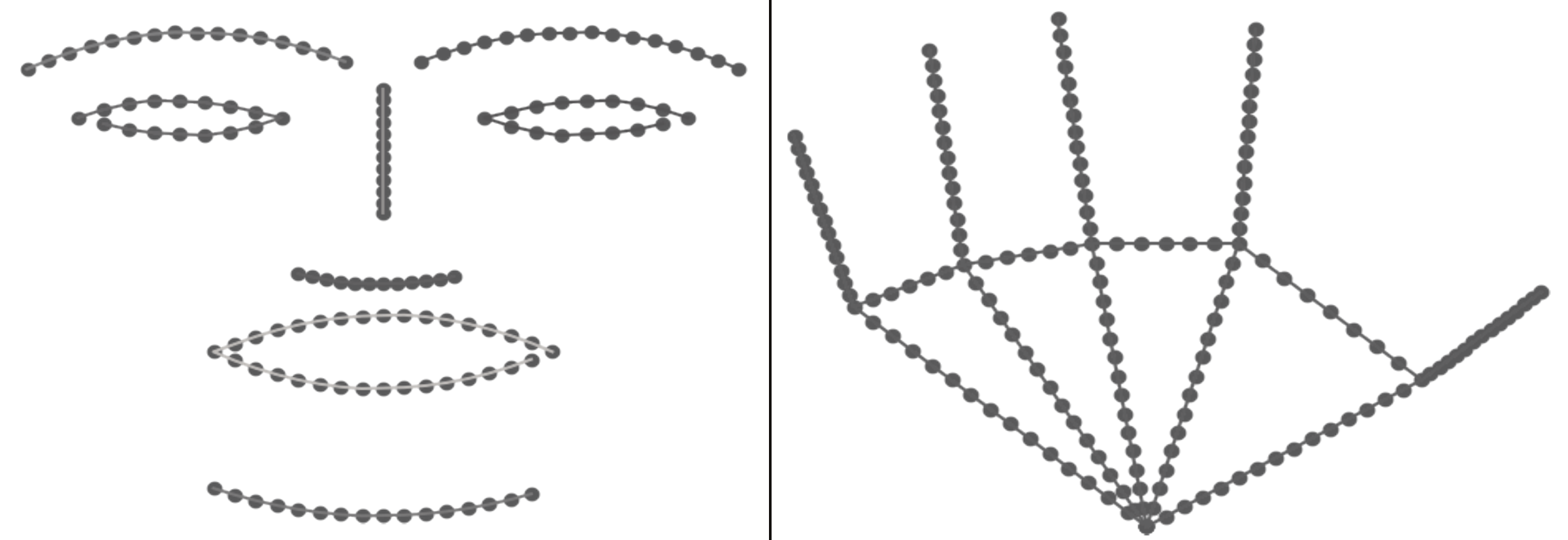}
  \vspace{-25px}
\end{wrapfigure}
Synthetically marked images resemble marked images while maintaining the spatial integrity of the template, where template landmarks on the unmarked images are intrinsically inferred. Our formulation is agnostic of the image class, and gives full control over the landmark definitions. Thus, we can even predict landmarks for body parts that previously lacked detailed training data, such as feet.
\vspace{4px}
\\
\textbf{Landmark Deformation Network $L_{\mathsf{\Delta}}$.} Similar to Zhu et. al ~\cite{zhu2017unpaired}, our network $L_{\mathsf{\Delta}}$ learns unmarked-to-marked images, where landmark deformation are used to simulate markers on images, Fig.~\ref{fig:ex1}. Cyclic consistency assures consistent learning. For landmark deformation prediction, we use a simple network that consists of 4 convolutional followed by 3 fully connected layers. During training, we employ an off-the-shelf generator network~\cite{zhu2017unpaired}, $G:\mathsf{I}^{\mathsf M}_{real} \rightarrow \mathsf{I}^{\mathsf U}_{fake}$ that translates marked images into fake unmarked images, and similarly discriminators $D^M$, $D^U$. When translating from domains $\mathsf{I}^{\mathsf U} \rightarrow \mathsf{I}^{\mathsf M}$, we first predict landmark template deformations, $\mathsf{\Delta} = L_{\mathsf{\Delta}}(\mathsf{I}^{\mathsf U}_{real})$ that are then rendered onto the input image via a differentiable renderer $R$, 
that gives us $\mathsf{I}^{\mathsf M}_{fake} = R(\mathsf{\Delta}, \mathsf{I}^{\mathsf U}_{real})$. 
\vspace{4px}
\\
\textbf{Spring Potential Loss.} In addition to CycleGAN loss above, we also employ a spring potential loss \cite{nealen2006physically}, that helps maintain spatial consistency of the initial semantic template. 
As a result, landmarks maintain their spatial structure w.r.t template definitions. This is essential for recovering smooth landmarks while regularizing the GAN loss. Since, we want these separations to be consistent with the original template, we define this loss in terms of the change in deformations between neighbouring pairs of landmarks, $\mathcal{L}_{spring}(t_i) = \mathsf{K}\Xsum_{t_j\in \{\mathsf{T}\}} || \mathsf{\Delta}_{ij}||_{2}^{2}$, where $\mathsf{K}$ defines the spring constant, and $\mathsf{\Delta}_{ij}$ the change in spring length for a neighbouring pair $\{t_i, t_j\}$ of template landmarks over which the spring loss is defined.  
Our full objective is given by:
\vspace{-5px}
\begin{align}
        &\mathcal{L}(L_\mathsf{\Delta}, G, D^U, D^M) = \mathcal{L}_{GAN}(L_\mathsf{\Delta},D^U,\mathsf{I}^{\mathsf M}_{real},\mathsf{I}^{\mathsf U}_{fake}) \nonumber\\
        &+ \mathcal{L}_{GAN}(G,D^M,\mathsf{I}^{\mathsf U}_{real},\mathsf{I}^{\mathsf M}_{fake}) + \mathcal{L}_{cyc}(L_\mathsf{\Delta},G) +\mathcal{L}_{spring}(L_{\Delta}; T) \nonumber
\end{align}
\vspace{-30px}

\section{Results}
\vspace{-5px}
The proposed method is capable of learning landmarks for various body regions, including faces and hands. Our training set consists of roughly 18k frames (about 10 minutes each) for each domain -- marked and unmarked, scaled down to resolution $128 \times 128$. The landmark template is rendered as 2D points via PyTorch3D \cite{ravi2020pytorch3d}. \textcolor{black}{Training took about 2 hours (wall-clock) on a RTX Titan.} Fig.~\ref{fig:results} shows results for face and hand body regions.
\vspace{-15px}
\section{Conclusion}
\vspace{-5px}
This paper presents a novel method for performing landmark prediction on body regions. While manual annotations suffer inconsistencies due to ambiguities of precise landmark locations, our method uses ground truth-like data for learning landmarks as template-based deformations that match the visible ground truth information when rendered, although in this work we do not model occluded marker regions. In addition, our method is not limited by body regions or landmarks for which there are no available datasets. 
\vspace{-13px}

\bibliographystyle{eg-alpha-doi}
\small\bibliography{00.main}

\end{document}